\documentclass[conference]{IEEEtran}
\IEEEoverridecommandlockouts
\usepackage{cite}
\usepackage{amsmath,amssymb,amsfonts}
\usepackage{graphicx}
\usepackage{textcomp}
\usepackage{xcolor}

\usepackage{algorithmicx}

\usepackage{algpseudocode}
\usepackage{subfigure}
\usepackage{times}
\usepackage[hyphens]{url}
\usepackage{multirow}
\usepackage{threeparttable}
\usepackage{supertabular}
\usepackage{booktabs}
\usepackage{epstopdf}
\usepackage{xspace}
\usepackage{enumitem}
\usepackage{array}

\usepackage{algorithm}

\newcommand{\MIMIC}{{\sc MIMIC III}\xspace}
\newcommand{\MedLens}{{\sc MedLens}\xspace}

\begin{document}


\title{MedLens: Improve Mortality Prediction Via Medical Signs Selecting and Regression}

\author{\IEEEauthorblockN{1\textsuperscript{st} Xuesong Ye$^*$}
\IEEEauthorblockA{
\textit{Trine University} \\
Phoenix, United States \\
xye221@my.trine.edu}
\and
\IEEEauthorblockN{2\textsuperscript{nd} Jun Wu}
\IEEEauthorblockA{
\textit{Georgia Institute of Technology} \\
Atlanta, United States \\
jwu772@gatech.edu}
\and
\IEEEauthorblockN{3\textsuperscript{rd} Chengjie Mou}
\IEEEauthorblockA{
\textit{Trine University} \\
Phoenix, United States \\
cmou22@my.trine.edu}
\and
\IEEEauthorblockN{4\textsuperscript{th} Weinan Dai}
\IEEEauthorblockA{
\textit{Trine University} \\
Phoenix, United States \\
wdai22@my.trine.edu}
}

\maketitle


\begin{abstract}
Monitoring the health status of patients and predicting mortality in advance is vital for providing patients with timely care and treatment. Massive medical signs in electronic health records (EHR) are fitted into advanced machine learning models to make predictions. However, the data-quality problem of original clinical signs is less discussed in the literature. Based on an in-depth measurement of the missing rate and correlation score across various medical signs and a large amount of patient hospital admission records, we discovered the comprehensive missing rate is extremely high, and a large number of useless signs could hurt the performance of prediction models. Then we concluded that only improving data-quality could improve the baseline accuracy of different prediction algorithms. We designed \MedLens, with an automatic vital medical signs selection approach via statistics and a flexible interpolation approach for high missing rate time series. After augmenting the data-quality of original medical signs, \MedLens applies ensemble classifiers to boost the accuracy and reduce the computation overhead at the same time. It achieves a very high accuracy performance of 0.96 AUC-ROC and 0.81 AUC-PR, which exceeds the previous benchmark.
\end{abstract}


\section{Introduction}
\label{sec:intro}

\subsection{Background and Motivation}
It is essential for doctors to accurately predict patient mortality in the intensive care unit (ICU), which enables them to create better treatment plans and risk assessments to increase patient survival. Manual analysis and interpretation of massive clinical data are time-consuming and rely on expert experiences. This is why machine learning is introduced for automatic analysis and intelligent decision-making. 

Most previous researches focus on using advanced machine learning or deep learning algorithms or optimizing algorithms to model clinical notes for accurate prediction. However, those works ignored to do measurements or characterizing original clinical notes. For example, the recording frequency is irregular, the missing rates of records are very large across most signals, and the amount of signal types is extremely manifold. The insufficient understanding of clinical data largely limits the performance of mortality prediction, even when using advanced artificial intelligence techniques.

After an in-depth measurement of a famous clinical dataset \MIMIC, we discovered several long-standing challenges of data quality and decided to solve them to improve the baseline accuracy of the mortality prediction task. We designed a generic pipeline called \MedLens, consisting of critical signals selection, a novel missing data interpolation approach, and ensemble time series classification. 

\subsection{Data Driven Health Analysis}
\subsubsection{Mortality Prediction}
\label{subsubsec:related_mortality}
Medical vital signs are essential measurements that indicate a person's body temperature, heart rate, and blood pressure, reflecting their overall health. In addition to medical signs, the most common records are clinical notes in textual format, written by physicians or nurses. A patient's health status strongly correlates with the varying patterns of medical signs. Several works solely focus on utilizing medical signs to predict mortality. Harutyunyan \emph{et al.} \cite{harutyunyan2019multitask} utilize multitask learning to predict mortality using clinical time series data, improving predictive performance. Che \emph{et al.} \cite{che2018recurrent} implement recurrent neural networks to handle missing values in multivariate time series, enhancing death prediction accuracy. Shukla \emph{et al.} \cite{shukla2021multi} employ multi-time attention networks to address irregularly sampled time series data, resulting in more accurate death predictions.

Many researchers have mined semantic information in clinical notes or combined it with medical signs to make more accurate mortality predictions. Deznabi \emph{et al.} \cite{deznabi2021predicting} combine clinical notes and time-series data to predict in-hospital mortality, resulting in enhanced prediction accuracy. Khadanga \emph{et al.} \cite{khadanga2019using} leverage clinical notes and time-series data to improve ICU management, leading to better patient outcomes and resource allocation. Some researchers focus on improving the quality of electronic health records (EHR) embeddings to enhance the performance of downstream prediction tasks. Zheng \emph{et al.} \cite{liu2020heterogeneous} investigate the relationship among various entities in EHR via a heterogeneous graph neural network. FineEHR \cite{wu2023fineehr} designed two text embedding approaches for label correlation and note categories, respectively, effectively improving the mortality prediction accuracy. Zheng \emph{et al.} \cite{liu2022mitigating} designed a re-weighting approach to balance modeling performance among different demographic groups.

\subsubsection{Auxiliary Medical Care}
Electronic health records are also useful in other important medical prediction tasks, such as diagnosing diseases and monitoring the health status. Some researchers focus on using medical signs in medical records for diagnosing diseases. For example, Park \emph{et al.} \cite{park2022deep} apply deep learning algorithms to analyze laboratory test data, enabling early detection of pancreatic cancer. Mehrdad \emph{et al.} \cite{mehrdad2022deterioration} utilize medical signs and clinical features to predict COVID-19 patient deterioration. In addition to medical signs, medical images also play an important role in diagnosing diseases. Yuli \emph{et al.} \cite{wang2023deep} propose a deep-learning architecture for retinal image registration. Ziyang \emph{et al.} \cite{wang2022adversarial} enhance the performance of Vision Transformer for medical image segmentation via adversarial training. Ziyang \emph{et al.} \cite{wang2023cnn} then explore the power of Vision Transformer in a semi-supervised situation for medical image segmentation. Furthermore, some researchers also utilize signs to monitor the health status of humans. DeepVS \cite{xie2022deepvs} propose a deep learning approach to use radio frequency signals for healthy status assessment. Zongxing \emph{et al.} \cite{xie2022passive} and \emph{et al.} \cite{xie2021signal} explore the use of non-touch signs such as Ultra-Wide Band (UWB), Wi-Fi, and millimeter wave to detect changes in healthy status and diseases for earlier health intervention.

\subsection{Time Series Prediction}
Using medical signs to predict mortality could be separated into two steps: \emph{time series interpolation} and \emph{time series classification}. The first target is to improve the data quality, especially when many values in a time series are missing (not recorded), and enhance the subsequent feature mining, modeling, and prediction tasks. The second step is finding the most suitable algorithm for modeling the correlation between time series features and the labels.

\subsubsection{Interpolation} 
\label{subsubsec:interpolation}
Interpolation is a comprehensive concept that targets filling in missing positions in a time series via various tactics. It is a common challenge in many time series analysis and prediction problems. In turn, many methods have been derived. Cheolhwan \emph{et al.} \cite{oh2020time} utilize interpolation methods, such as linear, polynomial, or spline, to fill gaps in time series data, creating augmented datasets for improved model training. Friedman \emph{et al.} \cite{friedman1962interpolation} employ related time series data for interpolation, leveraging correlation structures between series to estimate missing values in the target series.

Time series interpolation is frequently used to solve a variety of medical issues and is crucial for the pre-processing of data. Che \emph{et al.} \cite{che2018recurrent} apply recurrent neural networks to handle missing values in multivariate time series, enhancing death prediction accuracy. Luo \emph{et al.} \cite{luo2022evaluating} review various imputation methods for handling missing clinical data, assessing their impact on mortality prediction performance. Tipirneni \emph{et al.} \cite{tipirneni2022self} introduce a self-supervised transformer approach for sparse and irregularly sampled clinical time series, improving mortality prediction accuracy.

\subsubsection{Classification} 
Classification is the final step in many prediction tasks. Time series classification is widely investigated and applied to solve real-world problems. Traditional statistical approaches have the advantages of being quick and easy to build and deploy. Jeong \emph{et al.} \cite{jeong2011weighted} introduce a weighted dynamic time warping (DTW) method that adjusts the importance of time series elements, enhancing classification accuracy by considering local differences. Jun \emph{et al.} \cite{wu2023botshape} used clustering to analyze the similar patterns among time series and built an accurate time series classification system based on seasonality analysis and time series shape mining. Tahani \emph{et al.} \cite{daghistani2020comparison} compare the performance of the Random Forest Classifier and Logistic Regression algorithm on the task of diabetes prediction. CellPAD \cite{wu2018cellpad} introduce a novel Random Forest Regression-based approach for time series anomaly detection.

Currently, using deep learning models to predict or classify time series is becoming popular due to their flexible structure and ability to improve prediction performance. Karim \emph{et al.} \cite{karim2017lstm} combine LSTM and fully convolutional networks for time series classification, leveraging the strengths of both architectures in capturing long-term dependencies and spatial information. Karim \emph{et al.} \cite{karim2019multivariate} combine Long Short-Term Memory (LSTM) and Fully Convolutional Networks (FCNs) to create a multivariate LSTM-FCN architecture, effectively handling multivariate time series data for improved classification performance. Yifan \emph{et al.} \cite{li2020time} propose a GNN-based framework to optimize time series prediction by capturing both spatial and temporal dependencies. Zhao \emph{et al.} \cite{zhao2017convolutional} employ convolutional neural networks (CNNs) for time series classification, taking advantage of their ability to automatically learn discriminative features and handle large-scale data.

\section{Dataset}
\label{sec:dataset}

\begin{table}[h]
\centering
\caption{MOST SIGNIFICANT SIGNALS WITH STRONG CORRELATION WITH MORTALITY STATUS IN \MIMIC}
\label{tab:correlation} 
\vspace{-6pt}
\begin{tabular}{| c | c | p{1.8in} |}
\hline
{\bf Signs} & {\bf Pearson} & {\bf Description} \\
\hline
GCS Total & -0.305 & Score of consciousness by eye, verbal, and motor responses. Lower scores indicate reduced consciousness. \\
\hline
Motor Response & -0.283 & Response assesses limb movement in response to stimuli. Lower scores indicate reduced motor function. \\
\hline
Eye Opening & -0.277 & Both spontaneous and stimulus-induced eye opening. Lower scores indicate reduced responsiveness. \\
\hline
Verbal Response & -0.247 & Verbal response assesses verbal communication ability. Lower scores indicate reduced communication. \\
\hline
Urea Nitrogen & 0.202 & Urea nitrogen levels in blood evaluate kidney function. Higher levels suggest reduced kidney function.\\
\hline
RDW & 0.200 & RDW measures red blood cell size/shape variation. Higher values signal underlying health issues. \\
\hline
Heart Rate & -0.195 & Beats per minute. Lower values suggest reduced cardiovascular function.  \\
\hline
Platelet Count & -0.150 & Platelet count in blood. Lower counts indicate higher bleeding risk. \\
\hline
CVP & 0.141 & Central venous pressure (CVP) reflects blood return to the heart; higher values indicate venous pressure.  \\
\hline
pH & -0.134 & Blood pH measures acidity or alkalinity, imbalances may signal respiratory or metabolic issues. \\
\hline
\end{tabular}
\end{table}

Johnson \emph{et al.} \cite{johnson2016mimic} publishes a large database of ICU patient admission information in intensive care units(ICU). It contains clinical data recorded from 2001 and 2012, mainly including vital signs of patient health status and textual notes written by physicians, nurses, and medical technicians medications. Vital signs refer to laboratory measurements like blood pressure, heart rate, and blood PH. 

In each hospital admission record, the most important information is the final mortality status of the patient. \MIMIC accurately and clearly recorded it, which is a reliable ground-truth for mortality prediction task evaluation. In detail, in original admission records, the attribute HOSPITAL$\_$EXPIRE$\_$FLAG refers to the mortality status. If a patient dies during this hospitalization, the flag is set to 1. If not, the flag is 0. According to the statistics of \MIMIC, the total amount of admission records are hundreds of thousands, where the mortality is about one-tenth. 

The total amount of different sign categories reaches more than 7,000, which is very imbalanced across patients. To put it another way, a lot of signs are only recorded in a small group of patients. However, some vital signs are recorded almost for each patient. Table~\ref{tab:correlation} gives the top-10 significant signs with the strongest correlations (negative or positive) with the health status. Pearson refers to the Pearson Correlation Coefficient (detailed explanation in the section~\ref{subsec:measure_correlation}) between the values of the corresponding sign and the final mortality status in all admission cases.

\section{Measurement}
\label{sec:measure}

This section measures the data distribution of \MIMIC from the common data-quality views in data mining and machine learning fields. It targets to answer two questions:

(i) What low data-quality problems would limit the prediction performance?

(ii) How to alleviate their bad influences and mine effective information for prediction?

\subsection{Data-quality Analysis}
\subsubsection{Data missing and imbalance}

Data missing is a common problem for time series data because a complete time series should contain all timestamps with a fixed data collecting time interval during a period of time. However, a time series is hard to be completely recorded, especially in medical records. The reasons include the different life and work schedule of patients and physicians, also some extra events like surgery. 

\begin{figure}[h]
\centering
\includegraphics[width=3.5in]{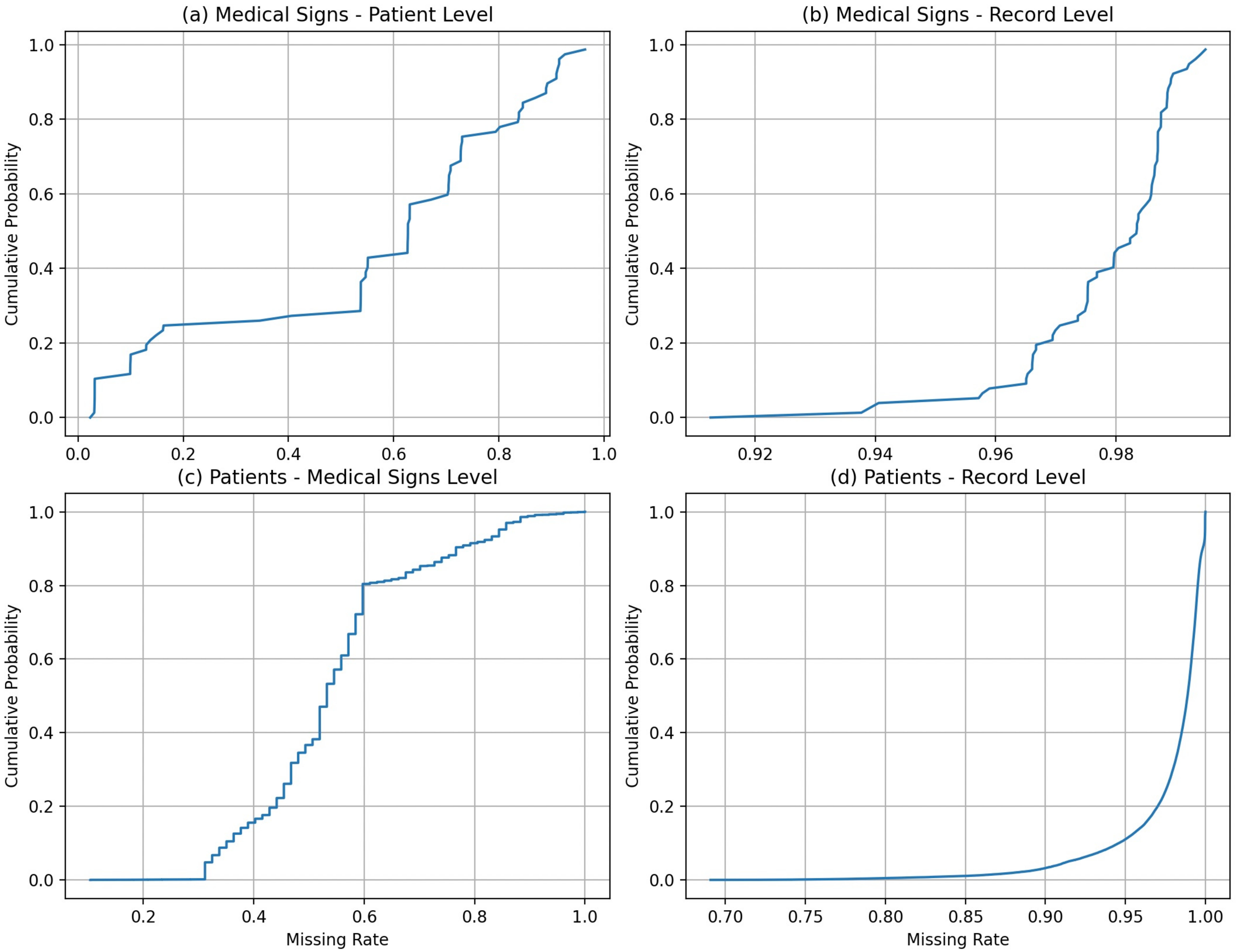}
\vspace{-6pt}
\caption{MIMIC-III data missing metrics}
\label{fig:missing_rate}
\end{figure}

In \MIMIC dataset, the missing rate of medical signs is comprehensively very high. To measure the missing data degree from all aspects, we designed four metrics and plotted their cumulative probability distribution in  Figure~\ref{fig:missing_rate} (We firstly select 77 types of medical signs with the highest recording frequency to reduce the computation):

\begin{itemize}
  \item (a) Medical Signs - Patient Level: We set the metric to measure the missing rates of various medical signs in patient level, equaling $1 - \frac{Count_P(i)}{Count_P}$, where $Count_P$ is the total count of patients and $Count_P(i)$ is the total count of patients having the records of the \emph{i-th} medical signs.
  \item (b) Medical Signs - Record Level:
  We set the metric to measure the missing rates of various medical signs in record level, equaling $1 - \frac{Count_R(i)}{Count_R}$, where $Count_R$ is the total count of hourly time series timestamps across all patient admissions, and $Count_R(i)$ is the total count of hours according to real records of the \emph{i-th} medical signs.
  \item (c) Patients - Medical Signs Level: Then we calculate the missing count of sign types for each patient, equaling $1 - \frac{\sum_{i=1}^{77}exsit(i)}{77}$.
  \item (d) Patients - Record Level: Similar to the setting in (c), we calculate the missing count of records across 77 sign types for each patient, equaling $1 - \frac{\sum_{i=1}^{77}{\sum_{t=1}^{48}exsit(i, t)}}{77}$.
\end{itemize}

According to the figures, we have several conclusions: (i) Medical signs tend to be rarely recorded every hour; (ii) Many medical signs are not recorded on all patients. They are also supported by the statistical fact that in all the numerical sign records, the top 77 medical sign types can account for 57.56\% of the total records. 

\subsubsection{Useless information}

\label{subsec:measure_correlation}
\begin{figure}[h]
\centering
\includegraphics[width=3.5in]{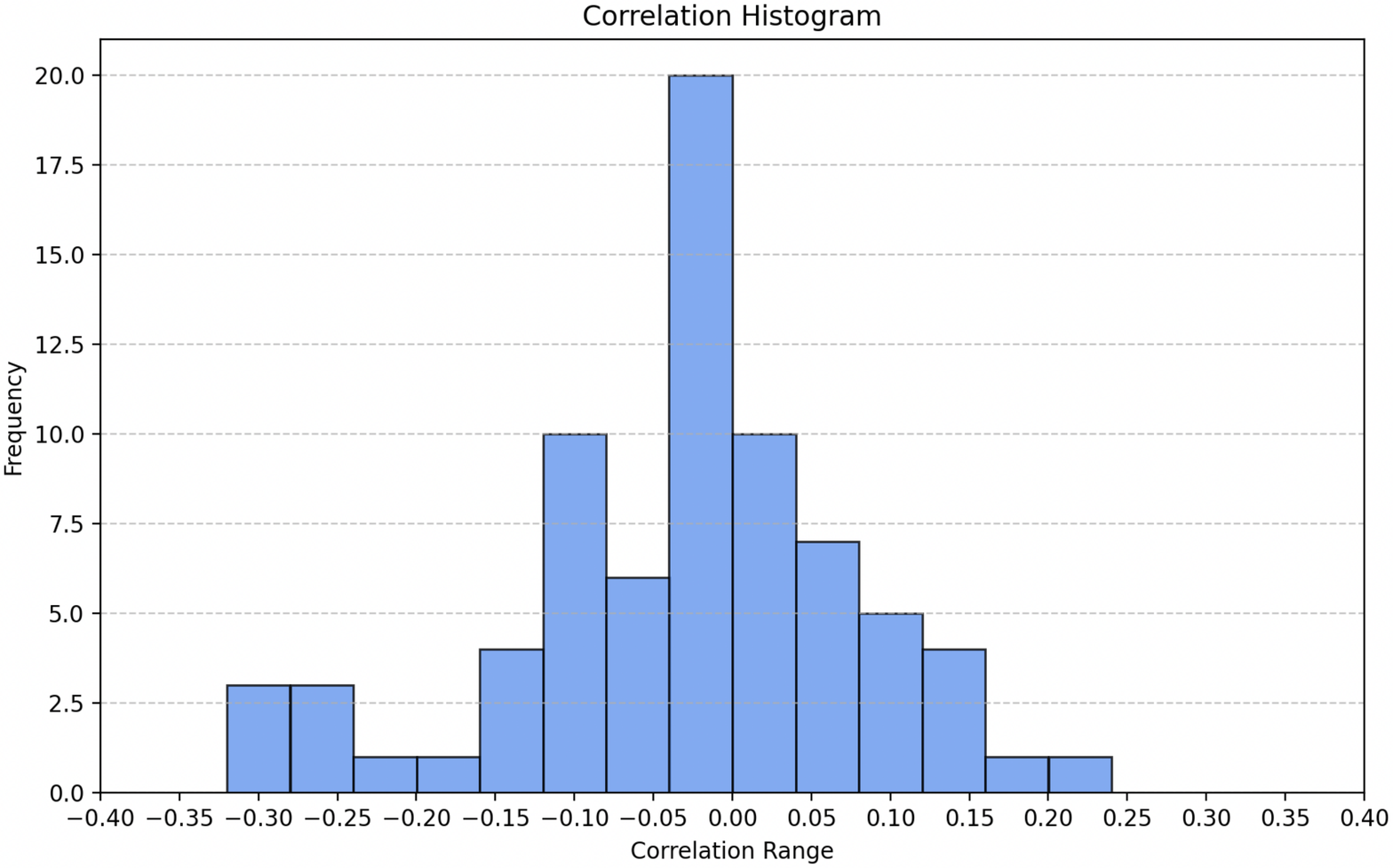}
\vspace{-6pt}
\caption{MIMIC-III correlation between medical signs and mortality status}
\label{fig:correlation_histogram}
\end{figure}

To better understand the correlation between the mortality label and each medical sign, specifically, we calculated the Pearson correlation coefficient, which measures the linear relationship between two variables for each medical sign and mortality record in \MIMIC. Figure~\ref{fig:correlation_histogram} show the histogram of Pearson scores across all medical signs. A higher absolute value of the correlation coefficient indicates a stronger linear relationship between the corresponding medical variables and the mortality label. The results that showed approximately 86\% medical signs are concentrated in the range between -0.15 and 0.15, with a low correlation with mortality status. The figure points out many signs are possibly useless and redundant information for mortality prediction.

\subsection{Bad Influences}
\label{subsec:measure_solution}
Missing positions in a time series set up obstacles for modeling because those time series is hard to do vector computing like feature generation and model fitting and predicting. Interpolation could alleviate this problem by predicting and filling in the most possible values on missing positions. However, a high missing rate also limits the interpolation accuracy because it could not provide enough information for time series analysis and pattern mining for interpolation. More features do not mean better prediction performance. On the contrary, useless features could become noise to the model and also increase the computation overhead. 

Based on the above hard situations, we need a simple but efficient mechanism to discover less but efficient medical signs for mortality prediction, and also need to design a proper interpolation algorithm to fill in missing data. Section~\ref{sec:design} introduces the detailed interpolation and mortality prediction approach.

\section{Design}
\label{sec:design}
\begin{figure}[htb]
\centering
\includegraphics[width=3.4in]{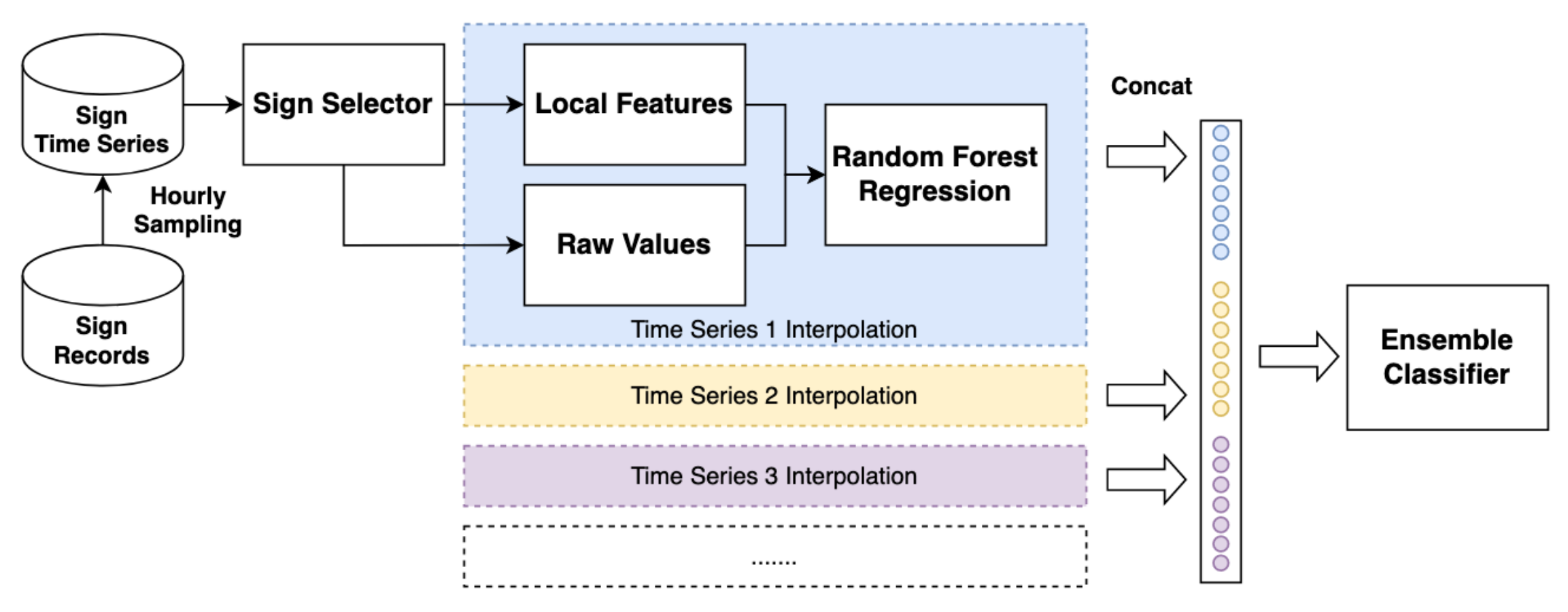}
\vspace{-6pt}
\caption{\MedLens Framework}
\label{fig:arch}
\end{figure}

\MedLens aims to improve the performance of mortality prediction by enhancing the information in raw medical signs. It takes electronic health records (EHR) as input and targets to predict whether the patient will die in the future. Figure~\ref{fig:arch} shows the architecture of the entire system.

\subsection{Hourly Sampling}
Original medical signs in \MIMIC are discontinuous records, which are the isolated timestamps and corresponding sign values. In order to generate fine-grained time series for a long period of time, \MedLens selects 1 hour as the time series interval and generates a 30-days time series for each sign and each patient admission. In detail, we will calculate the average value within each hour to obtain an hourly time series. A period of hospitalization records can be uniquely determined through the patient's ID and discharge time. The discharge time is rounded up by the hour and then pushed back 30 days. We calculate the mean value of the patient's medical variable data in each hour and assign the value to the right endpoint according to the principle of left opening and right closing.

\subsection{Sign Selector}
Before processing the data, we need to select the appropriate medical signs for analysis. In sub section~\ref{subsec:measure_solution}, we discussed the data-quality problems and the harm of useless information for the prediction model. \MedLens designed a simple but effective two-step feature selection approach to solving those problems. In detail, we first select a batch of medical signs with fewer missing rates (which could provide more information for time series interpolation and classification), then rank them according to their absolute values of Pearson correlation scores, and finally, filter out a subset of signs with top scores and correlation.

\subsection{Interpolation via Random Regression}
\begin{figure}[htb]
\centering
\includegraphics[width=3.2in]{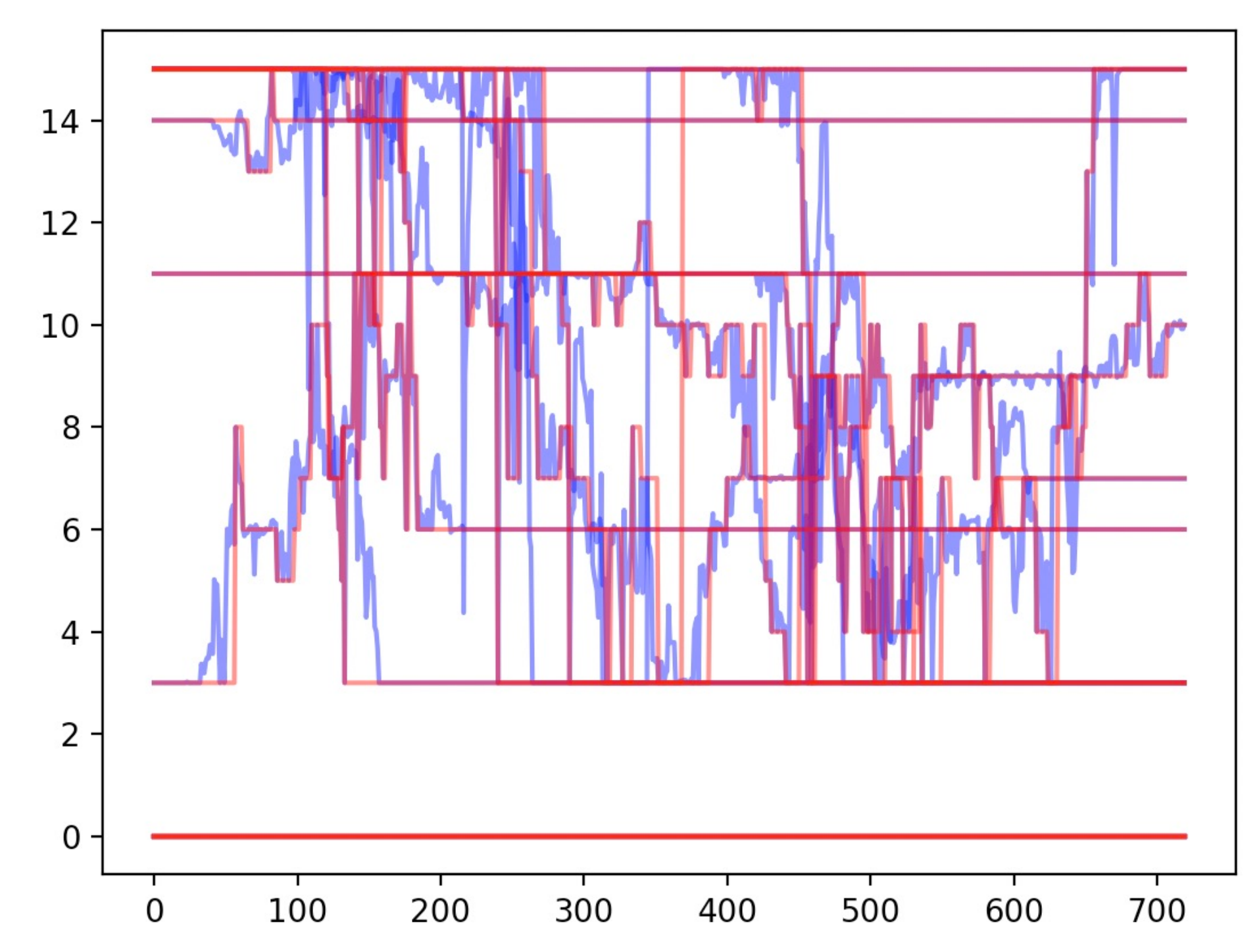}
\vspace{-6pt}
\caption{Compare Baseline and RF Interpolation}
\label{fig:interpolation}
\end{figure}
\label{sec:design}

\subsubsection{Algorithm Selection}
Different from mentioned interpolation approaches in section~\ref{subsubsec:interpolation}, \MedLens designs a regression approach to interpolate based on the consideration of large missing rate problems. Traditional pattern analysis and statistical approaches are not efficient in time series data with high missing rate, because they rely on continuous local sub-time series to make predictions. Deep learning approaches tend to utilize sequential information to achieve high accuracy. However, medical signs of humans are effected by time property. Under comprehensive consideration, we select a random forest regression algorithm to make interpolation because of two advantages:
(i) The instance generation process is flexible because it could throw away missing positions without enough local time series information.
(ii) The feature setting is flexible because it accepts nearly almost all numerical and categorical features.

\subsubsection{Feature Generation}
\MedLens uses the surrounding series of a missing position to construct the feature vector. It first sets a hyper-parameter $window$ to control the range of the surrounding series. Because the missing rate of local time series is various, we use a pooling approach to create features. In detail, our system will calculate the mean value, maximum value, minimum value, variation value, and standard deviation of the surrounding series, which are statistical metrics from various aspects to measure the local series. Besides, the time property is added to the vector feature by transforming the timestamp of the predicted position to the hour index.
\subsubsection{Interpolation}
\MedLens split the data into the training set and the prediction set. 

\emph{Training:} The predicted variable is the practical value at one time series time position. \MedLens drops illegal feature vectors or predicted values if any of them contains the missing value (NaN). And all remaining training samples are inputted to Random Forest Regression \cite{breiman2001random} for fitting the correlation between the feature vector and the value of time series at a specific time position based on a large number of training samples.

\emph{Interpolation:} 
On the test set, the feature generation process is the same, but the predicted variable is a missing value. In other words, the regressor uses the model and the feature vector to predict the value of a missing point and fill it in. 
Finally, \MedLens uses the forward filling and the backfilling to fill in remaining missing values (positions without an irregular feature vector), and finally replaces still remaining missing values with the default value zero (called \emph{baseline interpolation}).

Figure~\ref{fig:interpolation} shows the interpolation result, containing 20 randomly selected patients for a specific medical sign having the highest correlation score. The red lines are produced by the baseline interpolation approach, and the blue lines are produced by random regression interpolation. The blue lines are smoother, which shows trends among local time series.

\subsection{Mortality Prediction}
\label{subsubsec:medlens_mortality_prediction}
\MedLens applied the interpolated time series for mortality prediction. It first concats multiple interpolated time series into a unique vector for each patient according to the same sort of medical sign categories and sends it to a classifier as the \emph{feature vector}. Because the number of medical signs is large and the length of the time series is very long (30 days), \MedLens considers applying an ensemble classifier to boost the training and prediction speed and obtains high prediction accuracy at the same time. It utilizes the results from weak classifiers to vote and obtain the best-predicted label with high accuracy. It samples small batches of samples and features combination, which makes the overhead of computation to be very low and easy to speed up via parallelization. We consider the Random Forest classification algorithm  to be sufficient. The random forest improves the prediction accuracy by constructing and integrating multiple decision trees and realizes random sampling of features and samples. Compared to single decision trees, random forests are more robust and robust to noise and irrelevant features.

\section{Evaluation}
\label{sec:evaluation}

\subsection{Ground Truth}

\emph{Ground-truth} serves as a valuable means for gauging the predictive performance of labeled instances. We employed a ground-truth evaluation using the \MIMIC dataset to assess the efficacy of \MedLens. Mortality labels are indicated in the column \emph{HOSPITAL\_EXPIRE\_FLAG} within the admission records. The samples were then randomly split into two groups: the  \emph{training set} and the \emph{testing set}, with the latter group making up 20\% of the total.

\emph{Performance metrics} play a crucial role in comparing various models or methodologies. In this study, We considered the \emph{Accuracy} metric, which represents the proportion of correct predictions (both true positives and true negatives) among the total number of instances evaluated. While accuracy is an intuitive and easy-to-understand measure, it can sometimes be misleading in the presence of imbalanced datasets, as it tends to be biased towards the majority class. To address this limitation, we also employed the \emph{F1score}, which is the harmonic mean of precision and recall, aiming to strike a balance between these two metrics. A higher F1score indicates a better trade-off between precision (the proportion of true positive predictions among all positive predictions) and recall (the proportion of true positive predictions among all actual positive instances). This combination of accuracy and F1score allows for a more comprehensive evaluation of \MedLens performance, particularly in scenarios with imbalanced class distributions.

Additionally, we adopted AUC-ROC (the area under the receiver operating characteristic curve) and AUC-PR (the area under the precision recall curve) as our evaluation metrics. AUC-ROC represents a plot of the True Positive Rate (TPR) against the False Positive Rate (FPR). A higher AUC-ROC value signifies superior model performance in distinguishing between positive and negative classes across multiple threshold settings. A higher AUC-PR value indicates a more accurate prediction of mortality instances while reducing the occurrence of false alarms.

\subsection{Accuracy Performance and Time Consuming Across Various Classifiers}

\begin{table}[h]
\centering
\caption{Accuracy performance and training time consuming across various classifiers}
\label{tab:effective_increased_after_interpolation} 
\vspace{-6pt}
\begin{tabular}{| c | c | c | c |}
\hline
{\bf Methods} & {\bf Accuracy} & {\bf F1 Score} & {\bf Training Time} \\
\hline
K-Nearest Neighbors & 0.902 & 0.869 & 4 s \\
\hline
Logistic Regression & 0.931 & 0.926 & 9 s\\
\hline
Random Forest & 0.953 & 0.941 & 23 s \\
\hline
Multilayer Perceptron & 0.897 & 0.848 & 170 s \\
\hline
Gradient Boosting & 0.956 & 0.953 & 966 s \\
\hline
\end{tabular}
\end{table}

As seen in Table~\ref{tab:effective_increased_after_interpolation}, both ensemble algorithms, Random Forest and Gradient Boosting, outperform the other classifiers, MLP, KNN, and Logistic Regression in terms of accuracy and F1 score. This superior performance can be attributed to the inherent properties of ensemble classifiers, which combine the strengths of multiple weak learners to improve prediction accuracy and robustness. This is particularly important in the context of \MedLens, where a large number of medical signs and long-time series data are involved.

To ensure stable performance, we randomly selected 5000 records which were divided by discharge time. Accuracy and F1 Score in Table~\ref{tab:effective_increased_after_interpolation} are mean values after multiple samplings and predictions. In our experiments, which were conducted on a machine with a 6-core CPU and 16GB of RAM, Random Forest and Gradient Boosting demonstrated their ability to handle the complexity of the feature vectors produced by \MedLens, achieving high prediction accuracy while maintaining a manageable computational overhead. In comparison to Gradient Boosting, Random Forest exhibited a significantly faster computation time, consuming only 2.28\% of the time required by Gradient Boosting with only a slight decrease in accuracy. This advantage can be crucial when dealing with large-scale datasets, as it enables \MedLens to deliver high-quality predictions in a timely manner.

\subsection{Performance Under Different Interpolation Methods}

\begin{table}[h]
\centering
\caption{Performance under different interpolation methods}
\label{tab:compare_interpolation} 
\vspace{-6pt}
\begin{tabular}{| c | c | c | c | c | c |}
\hline
{\bf Interpolation} & {\bf Accuracy} & {\bf F1 Score} & {\bf AUC-ROC} & {\bf AUC-PR} \\
\hline
Baseline & 0.946 & 0.9405 &	0.9521 & 0.8042 \\
\hline
Random Forest & 0.953 & 0.9407 & 0.9553 & 0.8051 \\
\hline
\end{tabular}
\end{table}

In subsection~\ref{subsubsec:related_mortality}, we discussed some mortality prediction works via medical time series signs. Among them, Harutyunyan \emph{et al.} \cite{harutyunyan2019multitask} served as a benchmark for this task and designed a deep learning model for prediction. However, they only applied a baseline interpolation approach via forward filling and backward filling. We consider the lack of a proper interpolation process largely limits the accuracy of their model.

Therefore, we sought to improve upon the existing methods by employing a more sophisticated interpolation technique in our \MedLens system, aiming to provide a more accurate representation of medical time series data.

Table ~\ref{tab:compare_interpolation} showcases the improvement achieved using Random Forest interpolation over the Baseline method. Across all metrics, including Accuracy, F1 Score, AUC-ROC, and AUC-PR, Random Forest interpolation enhances performance, which further boosts the already strong baseline results across all evaluation metrics in the \MedLens system.

\subsection{Compared With Previous Works}

\begin{table}[h]
\centering
\caption{Compare with previous works}
\label{tab:comparison_previous_works} 
\vspace{-6pt}
\begin{tabular}{| c | c | c |}
\hline
{\bf Methods} & {\bf AUC-ROC} & {\bf AUC-PR}  \\
\hline
MedLen (Random Forest) & 0.9553 & 0.8051 \\
\hline
Harutyunyan \emph{et al.} \cite{harutyunyan2019multitask} & 0.87 & 0.533 \\
\hline
Che \emph{et al.} \cite{che2018recurrent} & 0.8529 & / \\
\hline
Shukla \emph{et al.} \cite{shukla2021multi}  & 0.8591 & / \\
\hline
\end{tabular}
\end{table}

To demonstrate the improvement of our system, we compared the AUC-ROC and AUC-PR of our method with those reported in previous works, as shown in Table~\ref{tab:comparison_previous_works}. The results indicate that the \MedLens system, utilizing the Random Forest classifier, significantly outperforms previous approaches, achieving an improvement in AUC-ROC of nearly 10\%.

This considerable enhancement in performance highlights the importance of addressing the challenges associated with handling missing values and irregularly sampled time series in the medical domain. By incorporating these improvements, our proposed \MedLens system provides a more effective solution for mortality prediction.

\section{Conclusions}
Most previous studies on mortality prediction focus on designing new machine learning algorithms but do not systematically address the issue of data quality, which can hinder the improvement of prediction performance. After thorough data measurement, we observed excessive low-frequency and low-correlated vital signs present in the raw data, which not only increase the burden of model training but also potentially harm prediction accuracy.

To address these challenges, we have designed a statistical method to rapidly identify effective features. We select high-frequency categories of medical signs that have a high record proportion in the dataset and discard a large number of long-tail and irrelevant features. Additionally, we exclude medical signs with weak correlation to the mortality label and evaluate their performance, demonstrating that only a small subset of high-frequency and high-correlated vital signs already achieves very high prediction accuracy.

Furthermore, we have investigated the benefits of employing a better time series interpolation approach to enhance mortality prediction performance. Our proposed method, \MedLens, utilizes a novel time series interpolation technique based on Random Forest Regression, which automatically learns time series regularities and fills missing values. This approach has shown performance improvements across all common evaluation metrics compared to common interpolation approaches. Moreover, \MedLens improves modeling speed and prediction accuracy by employing a lightweight ensemble classifier.

Comparing with previous works, \MedLens has demonstrated significant improvements in terms of both AUC-ROC and AUC-PR. These results support our original motivation that improving data quality can lead to a generic performance enhancement in mortality prediction tasks, even for other prediction tasks based on electronic health records data.


\bibliographystyle{IEEEtran}
\bibliography{paper}

\end{document}